\documentclass[10pt,twocolumn,letterpaper]{article}

\usepackage[algorithms]{wacv}      % To produce the REVIEW version for the algorithms track
\definecolor{wacvblue}{rgb}{0.21,0.49,0.74}
\usepackage[pagebackref,breaklinks,colorlinks,allcolors=wacvblue]{hyperref}
\usepackage{graphicx}
\def\wacvPaperID{24} % *** Enter the WACV Paper ID here
\def\confName{WACV}
\def\confYear{2027}

\title{PixelControl: Fine-Grained Condition Fidelity in Text-to-Image Diffusion}

\author{%
\vspace{0.5em}%
Xin Lin\textsuperscript{1}\quad
Haodong Li\textsuperscript{1}\quad 
Zhifei Zhang\textsuperscript{2}\quad
Yutong Yang\textsuperscript{1}\quad \\
Haitian Zheng\textsuperscript{2}\quad
Juanxi Tian\textsuperscript{3}\quad
Zhe Lin\textsuperscript{2}\quad
Truong Nguyen\textsuperscript{1}\quad \\
\vspace{0.5em}\small
\textsuperscript{1} University of California, San Diego\quad
\textsuperscript{2} Adobe Research\quad
\textsuperscript{3} Nanyang Technological University\quad
\\
}

\begin{document}

\twocolumn[{%
\renewcommand\twocolumn[1][]{#1}%
\vspace{-1em}
\maketitle
\centering
\vspace{-1em}
\includegraphics[width=0.99\linewidth]{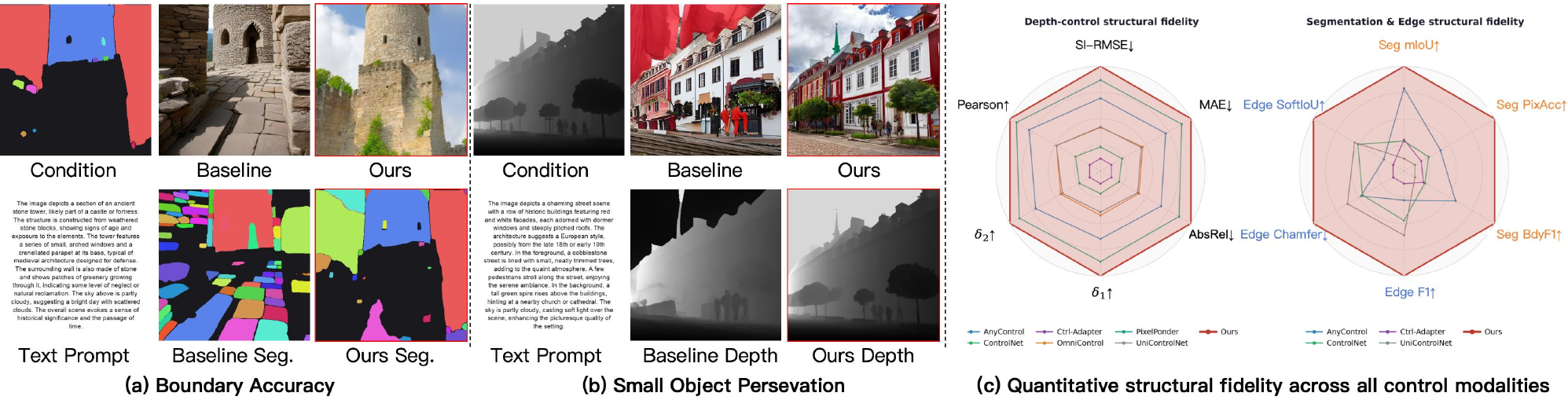}
\vspace{-1em}
\captionof{figure}{Motivation for fine-grained condition fidelity. 
(a) Existing methods can match coarse segmentation regions but drift around boundaries. 
(b) They can preserve global depth layout while omitting small conditioned regions. 
(c) Across depth, segmentation, and edge metrics, our PixelControl achieves stronger structural fidelity with high visual quality.
\vspace{1em}
}
\label{fig:teaser}
}]

\begin{abstract}
Controllable text-to-image diffusion models can often follow the global layout of spatial conditions, yet still violate fine-grained structures such as object boundaries, thin contours, and medium/small conditioned regions.
This limitation is especially problematic for VAE-based latent diffusion, where spatial compression can weaken high-frequency and low-area condition signals.
We propose PixelControl, a pixel-space controllable diffusion framework for fine-grained condition fidelity.
Built on a PixelDiT-style backbone, PixelControl avoids the latent bottleneck and introduces two complementary designs.
First, Structure-Aware Control Injection derives a condition structure map and uses it to strengthen injected control residuals around spatially sensitive regions.
Second, Multi-Scale Pyramid Cycle Loss verifies generated images against condition-derived structures across multiple resolutions, balancing global layout consistency with local boundary and detail accuracy.
PixelControl supports depth, segmentation, edge, and their combinations through modality-specific control branches with lightweight gated fusion.
Experiments across depth, segmentation, and edge control show that PixelControl improves structural fidelity and visual quality over existing controllable generation methods, with especially strong gains on boundaries and medium/small conditioned regions. The project page can be found at: \url{https://linxin0.github.io/pixelcontrol_homepage/pixelcontrol-site/}
\end{abstract}
    
\section{Introduction}
\label{sec:intro}

Controllable text-to-image generation aims to synthesize images that follow both a language prompt and an explicit spatial condition, such as depth, segmentation, edge, pose or layout.
Such spatial conditions control the structural distribution of the generated image, allowing users to specify not only what should appear but also where it should appear.
Recent controllable diffusion models~\cite{zhang2023controlnet,mou2024t2iadapter,zhao2024unicontrolnet,sun2024anycontrol,lin2025ctrl,tan2025ominicontrol,pan2025pixelponder} have made remarkable progress, which can follow the global shape and dominant contours of the condition while maintaining plausible image quality.

\iffalse
\begin{figure}[t]
  \centering
  \includegraphics[width=\columnwidth]{images/teaser.pdf}
  \caption{Motivation of fine-grained condition fidelity. Existing controllable generators can produce plausible images but still violate local conditions: boundaries and discontinuities may drift, and non-large conditioned structures may be weakened or omitted. Re-estimated controls are obtained from generated images using the same verifiers as evaluation, showing that PixelControl better preserves spatial alignment.}
  \label{fig:teaser}
\end{figure}
\fi

However, reasonable global layout does not imply accurate condition fidelity.
Figure \ref{fig:teaser} illustrates this gap with a closed-loop verification protocol.
We obtain input conditions from reference images using SAM2~\cite{ravi2024sam2} for segmentation and Depth Anything 3 (DA3)~\cite{lin2025depthanything3} for depth.
Given the same prompt and condition, each method generates an image, from which we re-estimate the corresponding structure using the same pretrained extractor~\cite{ravi2024sam2, lin2025depthanything3}.
Existing methods often preserve large structures and scene layouts, but fail to precisely match the condition at a finer granularity: for example, boundaries and local discontinuities may be spatially misaligned (Figure \ref{fig:teaser}(a)), while compact regions such as medium/small objects or thin structures may be weakened or omitted (Figure \ref{fig:teaser}(b)).
We summarize that this gap comes from two sources.
First, many controllable generators are built on VAE-based latent diffusion~\cite{rombach2022ldm}, where spatial compression and decoding can attenuate high-frequency condition cues, including sharp boundaries, local discontinuities, and low-area structures~\cite{medi2025missing,zheng2025diffusion}.
Second, existing control mechanisms usually inject and optimize control signals in a spatially uniform manner.
Large layouts, boundary regions, and compact components are treated with similar supervision despite requiring different levels of spatial precision.
Consequently, models tend to learn dominant global structures more reliably, while precise boundary alignment and compact condition details are easier to drift, blur, or disappear during denoising.

Recently, pixel diffusion~\cite{ma2026deco,ma2026pixelgen,yu2025pixeldit} has emerged as a promising alternative for high-fidelity image generation. Unlike latent diffusion models, pixel diffusion performs denoising directly in pixel space and avoids a VAE decoding bottleneck, making it better suited for preserving local structures and high-frequency contour details. Motivated by this property, we adopt a PixelDiT-style \cite{yu2025pixeldit} generator as the backbone for fine-grained controllable generation. 
Based on this observation, we design two complementary strategies to improve condition fidelity in the condition-injection process and the optimization objective, respectively.
At the injection level, we introduce Structure-Aware Control Injection (SACI), which derives a structure map from the condition and uses it to modulate the injected control residual.
Rather than applying the same control strength everywhere, SACI adaptively emphasizes spatially sensitive regions, including sharp boundaries, local discontinuities, and compact components.
This gives high-precision regions stronger influence during denoising, reducing boundary drift and the omission of fine condition details.
At the optimization level, we propose Multi-Scale Pyramid Cycle Loss (MPCL), which verifies the generated image against the condition at multiple resolutions.
Coarse scales regularize global layout consistency, while high-resolution scales impose stricter supervision on local alignment and fine structures.
Together, SACI strengthens condition guidance during generation, whereas MPCL provides scale-aware feedback during training.

The framework supports depth, segmentation, edge, and their combinations through modality-specific branches and a lightweight gating network over active controls.
While this design enables compositional control, multi-condition fusion is not our central claim.
As shown in \cref{fig:teaser}(c), extensive experiments in depth, segmentation, and edge conditions show that PixelControl achieves stronger condition fidelity while maintaining high visual quality.

The contributions of this paper are:
\begin{itemize}
  \item We propose PixelControl, a PixelDiT-based controllable generation framework for fine-grained fidelity. Extensive experiments on depth, segmentation, and edge conditions show that PixelControl improves structural fidelity and visual quality over existing controllable generators.
  \item We introduce Structure-Aware Control Injection, which derives a condition-aware structure map and strengthens residuals around spatially sensitive regions, improving boundary alignment and local condition preservation.
  \item We present Multi-Scale Pyramid Cycle Loss, which verifies generated images against condition-derived structures at multiple resolutions, providing scale-aware supervision from global layout to fine local alignment.
\end{itemize}

\section{Related Work}
\label{sec:related}

\noindent\textbf{Controllable text-to-image diffusion.}
Controllable text-to-image generation extends prompt-based synthesis with explicit spatial conditions, such as depth, edges, segmentation, pose, boxes, and layouts.
Existing methods inject external controls through trainable branches, lightweight adapters, grounded attention, or unified multi-condition interfaces~\cite{zhang2023controlnet,mou2024t2iadapter,li2023gligen,zhao2024unicontrolnet,sun2024anycontrol,lin2025ctrl,tan2025ominicontrol}.
Other studies train compositional or unified controllable generators to handle diverse spatial and semantic conditions within a single framework~\cite{huang2023composer,qin2023unicontrol,li2024controlnet++}.
Recent work also explores adaptive controllable generation~\cite{pan2025pixelponder}.

\noindent\textbf{Pixel Diffusion Models.} Pixel diffusion models provide an alternative route by performing generation directly in pixel space.
Early pixel-space diffusion and cascaded diffusion models demonstrate strong image synthesis and super-resolution capability~\cite{ho2020ddpm,dhariwal2021diffusion,ho2022cascaded,saharia2022imagen}.
Simple Diffusion~\cite{hoogeboom2023simple} further shows that high-resolution pixel diffusion can be effective with careful design, while recent transformer denoisers such as PixelGen, JiT and PixelDiT show the scalability of token-based diffusion backbones~\cite{ma2026pixelgen, li2026back, yu2025pixeldit}.

\noindent\textbf{Fine-grained and region-aware fidelity.}
Accurate condition following requires more than matching global layouts: boundaries should be spatially aligned, local discontinuities should be preserved, and compact condition regions should not disappear.
Multi-scale representations are widely used in vision to balance semantic context and localization accuracy.
U-Net~\cite{ronneberger2015unet} combines coarse features with high-resolution localization, while FPN~\cite{lin2017fpn}, Mask R-CNN~\cite{he2017maskrcnn}, PANet~\cite{liu2018panet}, and COCO-style area-binned evaluation~\cite{lin2014coco} emphasize the importance of scale-aware representation and assessment.
Frequency- and boundary-aware objectives also show that uniform image losses can under-emphasize hard high-frequency or contour regions~\cite{jiang2021focalfreq,kervadec2019boundary,yuan2020segfix}.

\section{Method}
\label{sec:method}

\begin{figure*}[t]
  \centering
  \includegraphics[width=\textwidth]{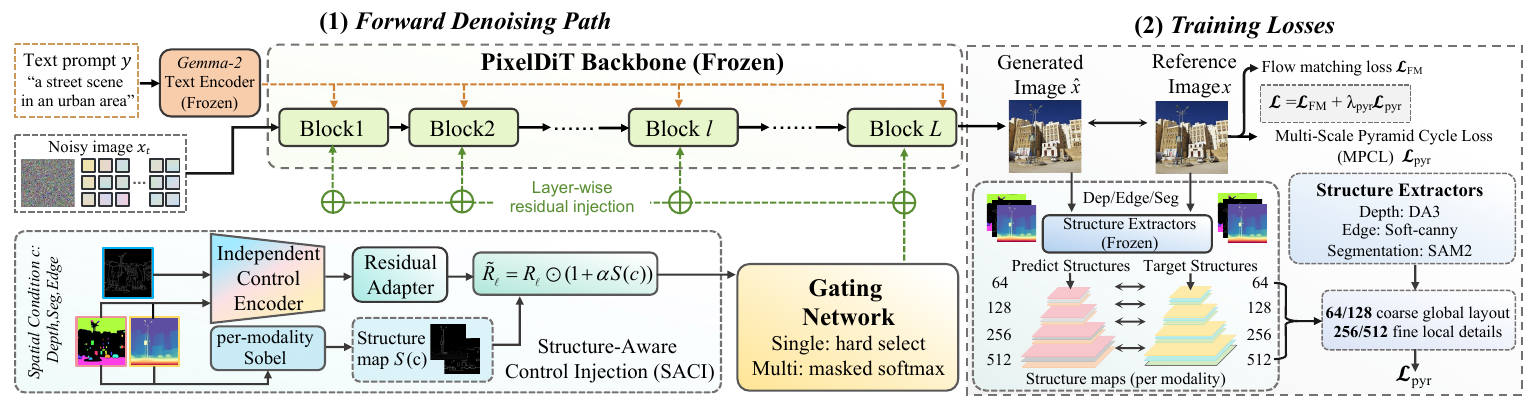}
  \caption{Overview of the proposed PixelControl. The condition is injected into the backbone through structure-aware residual modulation, and the model is trained with flow matching and multi-scale cycle supervision to preserve both global layout and fine condition details.}
  \label{fig:overview}
\end{figure*}

\subsection{Overview}
\label{sec:overview}

Given a text prompt \(y\), a noisy image \(x_t\), and a spatial condition \(c\), our goal is to generate an image that is not only globally plausible but also accurately aligned with the condition in fine regions. As shown in Figure \ref{fig:overview}, we build on a PixelDiT-style flow denoiser \(v_{\theta}\), which predicts the flow-matching velocity directly in pixel space. The condition is encoded into control tokens and injected into $L$=14 PixelDiT blocks through residual adapters. To make the injected signal more attentive to condition details, Structure-Aware Control Injection (SACI, Sec. \ref{sec:injection}) derives a structure map from the condition and amplifies residuals around compact components and discontinuities. During training, Multi-Scale Pyramid Cycle Loss (MPCL, Sec. \ref{sec:pyramid}) compares generated and condition-derived structures at multiple resolutions, using coarse scales for global layout consistency and high-resolution scales for fine boundaries and local structures. A lightweight layer-wise gating network selects the active control branch for single-condition inputs and balances multiple active branches for multi-condition inputs.

\begin{figure}[t]
  \centering
  \includegraphics[width=\columnwidth]{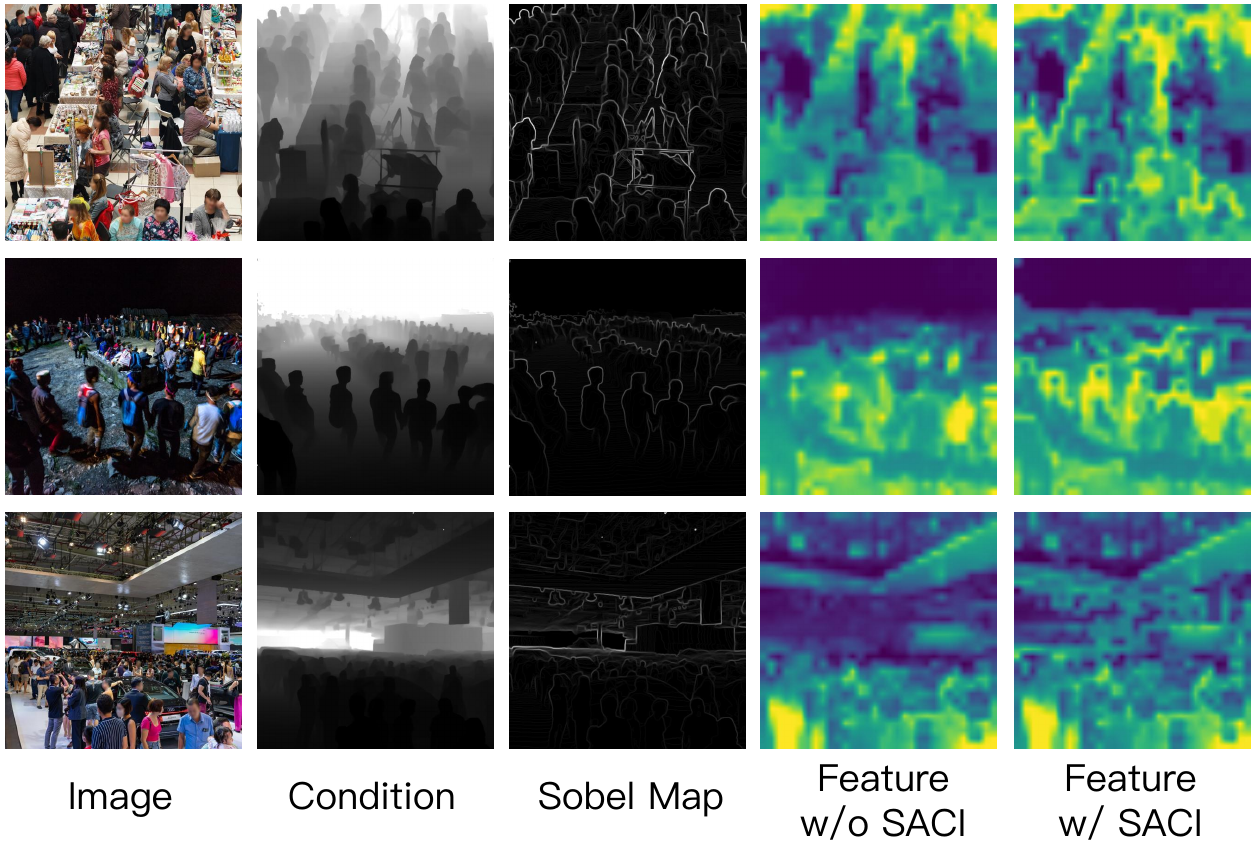}
    \caption{Effect of SACI on injected control features. From left to right, we show the original image, the input condition, the Sobel structure map derived from the condition, the injected feature without SACI, and the structure-modulated feature with SACI. SACI strengthens the injected control response around boundaries and local discontinuities.}
  \label{fig:structure}
\end{figure}

\begin{table*}[t]
  \centering
  \small
  \begin{tabular*}{\textwidth}{@{\extracolsep{\fill}}lccccccc@{}}
    \toprule
    Method & SI-RMSE \(\downarrow\) & MAE \(\downarrow\) & AbsRel \(\downarrow\) & \(\delta_1\uparrow\) & \(\delta_2\uparrow\) & \(\delta_3\uparrow\) & Pearson \(\uparrow\) \\
    \midrule
    AnyControl & 0.1277 & 0.0908 & 0.2715 & 0.6241 & 0.8215 & 0.9112 & 0.8310 \\
    ControlNet & 0.2103 & 0.1680 & 0.4514 & 0.3987 & 0.6241 & 0.7752 & 0.5593 \\
    Ctrl-Adapter & 0.2300 & 0.1869 & 0.4842 & 0.3490 & 0.5780 & 0.7482 & 0.4762 \\
    OminiControl & 0.1774 & 0.1317 & 0.3658 & 0.5067 & 0.7086 & 0.8291 & 0.6709 \\
    PixelPonder & 0.0971 & 0.0625 & 0.1949 & 0.7374 & 0.8956 & 0.9507 & 0.9049 \\
    UniControlNet & 0.1764 & 0.1340 & 0.3719 & 0.4903 & 0.7011 & 0.8257 & 0.6713 \\
    Ours & \textbf{0.0738} & \textbf{0.0460} & \textbf{0.1434} & \textbf{0.8081} & \textbf{0.9375} & \textbf{0.9744} & \textbf{0.9432} \\
    \bottomrule
  \end{tabular*}
  \caption{Depth-control structural fidelity. Best results are bolded.}
  \label{tab:depth}
\end{table*}

\subsection{Structure-aware control injection}
\label{sec:injection}

Standard residual control treats all condition locations equally. This is insufficient for fine-grained fidelity because different regions require different levels of spatial precision: smooth interiors mainly determine coarse layout, while boundaries and local discontinuities determine whether the generated image is accurately aligned with the condition. We therefore modulate the injected residual by a condition-derived structure map.

For a condition \(c\), the control encoder \(E\) produces condition tokens. At injection layer \(\ell\), a zero-initialized adapter maps these tokens to a residual:
\begin{equation}
  R_{\ell}= \gamma_{\ell} W_{\ell}\mathrm{LN}(E(c)),
  \label{eq:adapter}
\end{equation}
where \(W_{\ell}\) is zero-initialized and \(\gamma_{\ell}\) is a learnable scalar gate, so the pretrained denoiser is preserved at the start of training.
To make this residual spatially aware, we compute a structure map \(S(c)\) from the Sobel gradient magnitude of the condition. The response is normalized within each sample and resized to the token grid of layer \(\ell\). Intuitively, \(S(c)\) highlights spatially sensitive locations where the condition changes rapidly, such as segmentation boundaries and depth discontinuities. These regions require more precise alignment than smooth interiors, but uniform residual injection gives them the same strength as all other locations. Examples are shown in Figure \ref{fig:structure}. The injected residual is then
\begin{equation}
  \tilde{R}_{\ell}=R_{\ell}\odot(1+\alpha S(c)).
  \label{eq:structure}
\end{equation}
This modulation strengthens control around boundaries and local discontinuities without changing the pretrained backbone, improving spatial precision where condition fidelity is most sensitive to drift. The operation is spatially selective but parameter-free once the structure map is computed, so it adds little overhead and can be applied at every injection layer. 

We apply SACI only to depth and segmentation controls and enable structure injection with \(\alpha=2.0\): their dense or region-level conditions contain complete structural layouts, whose gradients provide reliable boundary weights.
We do not use the provided Canny conditions directly as they are sparse, threshold-dependent, and may fragment structures after RGB blur and binarization.
For edge control, SACI is disabled since the condition itself already encodes high-frequency boundaries, making additional edge amplification redundant and prone to noisy contours.

\subsection{Multi-scale pyramid cycle loss}
\label{sec:pyramid}

Condition fidelity should be verified at multiple spatial scales because different resolutions encode complementary structural cues.
Multi-scale designs in segmentation and detection show that coarse representations provide semantic context and large receptive fields, whereas high-resolution representations preserve localization signals for boundaries and small objects~\cite{ronneberger2015unet,lin2017fpn,liu2018panet,lin2014coco}.
Following this principle, low-resolution comparisons supervise global layout and large object geometry, while high-resolution comparisons impose stricter constraints on boundary drift, local discontinuities, and small spatial structures.
We therefore use a pyramid cycle objective to provide scale-aware feedback from coarse layout to fine local alignment.

Let \(\hat{x}\) be the generated image and \(x\) the reference image.
We first apply a modality-specific structure verifier \(\Phi_m\) to both images:
\begin{equation}
  \hat{s}^{m}=\Phi_m(\hat{x}), \qquad
  s^{m}=\Phi_m(x),
  \label{eq:verify}
\end{equation}
where \(m\in\{\mathrm{depth},\mathrm{seg},\mathrm{edge}\}\) denotes the active condition modality, and \(\hat{s}^{m}\) and \(s^{m}\) are the predicted and target structural maps.
For depth, \(\Phi_m\) is a frozen depth estimator; for segmentation, it is a frozen segmentation model; and for edge, it is a differentiable SoftCanny operator with a shared sampled threshold \(\tau\sim U(70/255,150/255)\) for generated and reference images.
The target branch is detached from gradients.

We then build a fixed resolution pyramid over the verified structural maps:
\begin{equation}
  \mathcal{P}=\{512,256,128,64\}, \qquad
  \boldsymbol{\beta}=(0.75,0.5,0.5,0.25),
  \label{eq:pyr_scales}
\end{equation}
where \(\beta_r\) denotes the loss weight for scale \(r\).
For each scale \(r\in\mathcal{P}\), the structural maps are resized as
\begin{equation}
  \hat{s}^{m}_r=\mathcal{D}_r(\hat{s}^{m}), \qquad
  s^{m}_r=\mathcal{D}_r(s^{m}),
  \label{eq:pyr_resize}
\end{equation}
where \(\mathcal{D}_r(\cdot)\) denotes resizing to resolution \(r\times r\).

The pyramid cycle loss is defined as
\begin{equation}
  \mathcal{L}_{\mathrm{pyr}}^{m}=
  \sum_{r\in\{512,256,128,64\}}
  \beta_r\,
  d\left(
  \hat{s}^{m}_r,
  \mathrm{sg}\!\left[s^{m}_r\right]
  \right),
  \label{eq:pyr}
\end{equation}
where \(d\) is SmoothL1 and \(\mathrm{sg}\) stops gradients through the target.
In this pyramid, coarse scales such as \(64\) and \(128\) constrain the overall structural distribution, while finer scales such as \(256\) and \(512\) impose stricter penalties on boundary misalignment, local discontinuities, and small spatial structures.
For edge control, using the same sampled SoftCanny extractor for both images avoids direct regression to threshold-sensitive offline Canny labels and makes supervision depend on structural consistency rather than a fixed preprocessing artifact.

The final training objective adds the proposed pyramid cycle supervision to the standard flow-matching loss:
\begin{equation}
  \mathcal{L}=
  \mathcal{L}_{\mathrm{FM}}
  +\lambda_{\mathrm{pyr}}\mathcal{L}_{\mathrm{pyr}}^{m},
  \label{eq:total}
\end{equation}
where \(\mathcal{L}_{\mathrm{FM}}\) is the original PixelDiT flow-matching objective and \(\lambda_{\mathrm{pyr}}=0.005\).

\section{Experiments}
\label{sec:experiments}

\begin{figure*}[t]
  \centering
  \includegraphics[width=\textwidth]{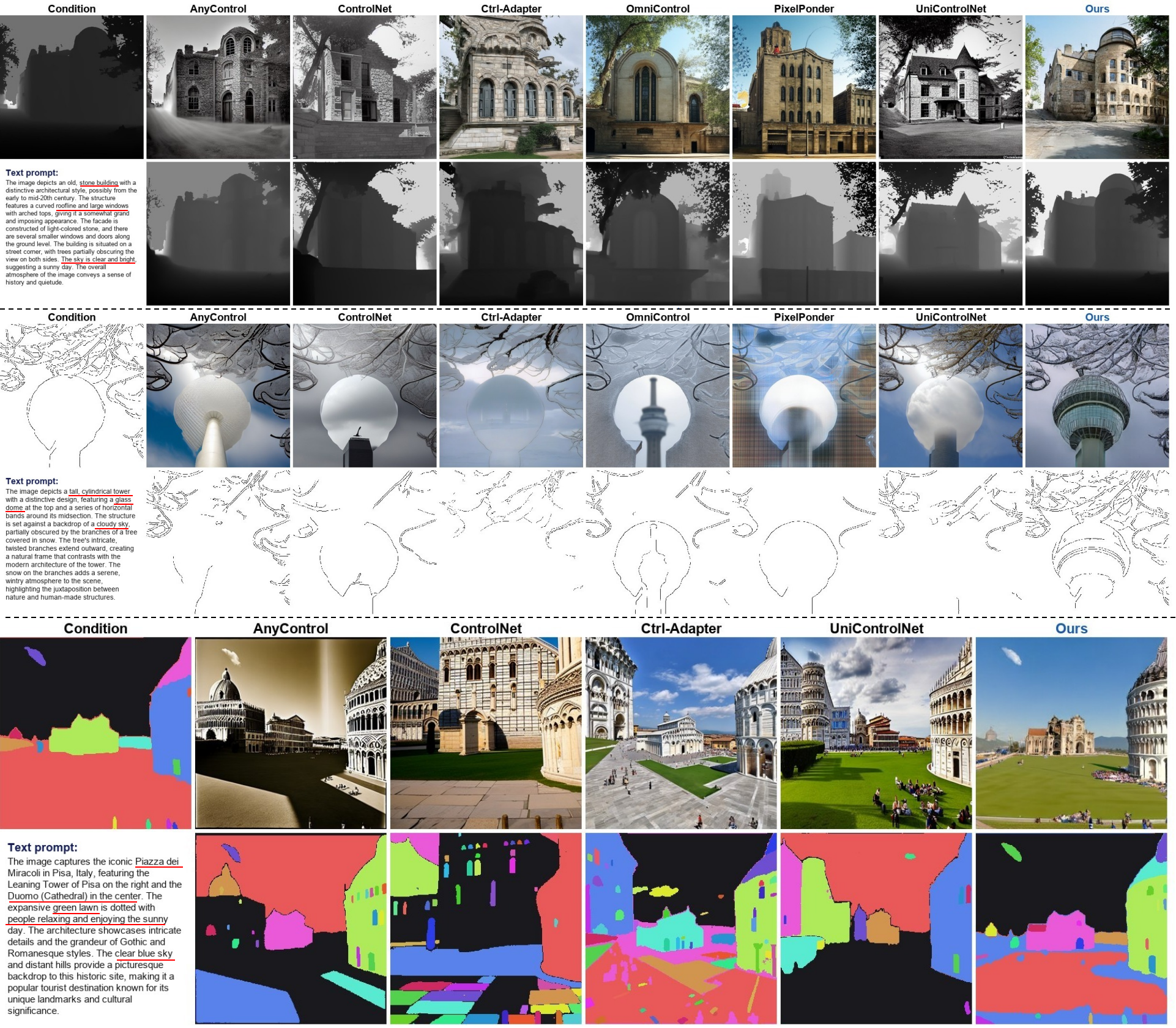}
  \caption{Single-condition qualitative comparison across depth, edge, and segmentation controls. For each example, we show the input condition, generated images from all methods, and the re-estimated maps.}
  \label{fig:qual}
\end{figure*}

\subsection{Experimental setting}
\label{sec:setup}

\noindent\textbf{Implementation details.}
We use PixelDiT-T2I~\cite{yu2025pixeldit} as the base pixel-space text-to-image generator and load its 1.3B pretrained model. Following the original PixelDiT setting, text conditioning is provided by the pretrained Gemma-2 \cite{team2024gemma} text encoder and fused through the MM-DiT \cite{esser2024scaling} conditioning pathway. Images are generated at \(512\times512\) resolution. The control pathway uses hidden dimension 1536 and injects residuals into 14 PixelDiT blocks. 
All these parts are kept frozen during control training, only the condition encoders, residual adapters, and lightweight multi-condition gating network are optimized. 
We train the model on NVIDIA A100 GPUs with a learning rate of \(1.0\times10^{-5}\).
Inference follows PixelDiT-T2I and uses FlowDPMSolver with 50 denoising steps, classifier-free guidance scale 2.75, and a batch size of 8 per GPU.

\noindent\textbf{Dataset and benchmark.}
We build our training set from the BLIP3o pretraining image-caption collection~\cite{chen2025blip3o}, which contains large-scale open-domain images paired with detailed long captions generated by Qwen2.5-VL \cite{wu2025qwen}.
The dataset covers diverse scenes, object categories, human activities, indoor/outdoor environments, and rich spatial compositions, making it suitable for evaluating controllable generation across both large layouts and medium/small objects.
We sample 2M image-caption pairs for training and reserve 2,000 images from a held-out shard for evaluation. 
For each RGB-caption pair, we construct depth, segmentation, and edge conditions using Depth Anything 3 (DA3)~\cite{lin2025depthanything3}, SAM2~\cite{ravi2024sam2}, and Canny edge detection~\cite{canny1986computational}, respectively.

Beyond full-image evaluation, we build a region-aware evaluation to measure fine-grained condition fidelity. 
We run YOLO26~\cite{yolo26} on each reference image to obtain method-independent object boxes, and divide them by relative box area \(a=|B|/(HW)\): small objects satisfy \(a<0.02\), medium objects satisfy \(0.02\leq a<0.10\), and large objects satisfy \(a\geq0.10\). 
We report the non-large bucket, defined as small and medium objects with \(a<0.10\). 
To ensure that each evaluated object is actually specified by the control signal, we further apply a condition-grounding gate: segmentation boxes must overlap a condition component, edge boxes must contain sufficient condition edge pixels, and depth boxes are kept directly because depth is dense. 
This protocol evaluates only YOLO-detected non-large objects that are grounded in the input condition, exposing local fidelity failures that are often hidden by whole-image metrics dominated by background and large foreground regions.

\begin{table}[t]
  \centering
  \resizebox{\columnwidth}{!}{
  \begin{tabular}{@{}lcccc@{}}
    \toprule
    Method & mIoU \(\uparrow\) & Pixel Acc. \(\uparrow\) & mAcc \(\uparrow\) & Boundary-F1 \(\uparrow\) \\
    \midrule
    AnyControl & 0.5220 & 0.7204 & 0.6006 & 0.5061 \\
    ControlNet & 0.3496 & 0.6955 & 0.5065 & 0.3558 \\
    Ctrl-Adapter & 0.3544 & 0.6797 & 0.4747 & 0.3488 \\
    UniControlNet & 0.2922 & 0.6664 & 0.4184 & 0.3070 \\
    Ours & \textbf{0.5943} & \textbf{0.8352} & \textbf{0.6577} & \textbf{0.6973} \\
    \bottomrule
  \end{tabular}}
  \caption{Segmentation-control structural fidelity. Best results are bolded.}
  \label{tab:segstructure}
\end{table}

\begin{table}[t]
  \centering
  \resizebox{\columnwidth}{!}{
  \begin{tabular}{@{}lcccc@{}}
    \toprule
    Method & Edge F1 \(\uparrow\) & Recall \(\uparrow\) & Chamfer \(\downarrow\) & Soft-IoU \(\uparrow\) \\
    \midrule
    AnyControl & 0.4127 & 0.4195 & 13.9251 & 0.0595 \\
    ControlNet & 0.4927 & 0.5051 & 14.1997 & 0.1256 \\
    Ctrl-Adapter & 0.3472 & 0.2633 & 19.6873 & 0.0355 \\
    OminiControl & 0.4850 & 0.3755 & 17.8455 & 0.1734 \\
    PixelPonder & 0.3211 & 0.2281 & 33.3204 & 0.1025 \\
    UniControlNet & 0.5517 & 0.6163 & 11.5315 & 0.1341 \\
    Ours & \textbf{0.7121} & \textbf{0.7163} & \textbf{5.4560} & \textbf{0.2387} \\
    \bottomrule
  \end{tabular}}
  \caption{Edge-control structural fidelity. Best results are bolded.}
  \label{tab:edgestructure}
\end{table}

\noindent\textbf{Compared methods.}
Since several recent methods have not released code, we compare with controllable text-to-image methods for which public implementations and pretrained models are available: ControlNet~\cite{zhang2023controlnet}, AnyControl~\cite{sun2024anycontrol}, Ctrl-Adapter~\cite{lin2025ctrl}, UniControlNet~\cite{zhao2024unicontrolnet}, OminiControl~\cite{tan2025ominicontrol}, and PixelPonder~\cite{pan2025pixelponder}. 
Both OminiControl and PixelPonder are based on the 12B FLUX.1 \cite{labs2024flux} model. Some baselines do not provide all condition types; these entries are marked with dashes in the tables.

\subsection{Comparison with existing methods}
\label{sec:comparison}

\noindent\textbf{Control fidelity.}
We first evaluate whether the generated images accurately preserve the input condition.
For fairness, each generated image is re-processed with the same structure extractor used to build the corresponding dataset condition.
For depth control, we re-estimate depth using Depth Anything 3~\cite{lin2025depthanything3} and compare it with the input depth condition using scale-invariant RMSE, MAE, AbsRel, threshold accuracies, and Pearson correlation.
For segmentation control, we re-segment generated images using SAM2~\cite{ravi2024sam2} and compare the predicted segmentation with the input segmentation condition, reporting mIoU, pixel accuracy, mean accuracy, and Boundary-F1.
For edge control, we extract edges using the same Canny-based pipeline~\cite{canny1986computational} as the condition construction, and report Edge F1, recall, Chamfer distance, and Soft-IoU.

As shown in \cref{tab:depth,tab:segstructure,tab:edgestructure}, PixelControl consistently improves structural fidelity across all three types of conditions. On depth, it reduces AbsRel from 0.1949 for the strongest baseline to 0.1434 and improves Pearson correlation to 0.9432, indicating that the generated scene geometry remains closer to the control map. On segmentation, the largest gain appears on Boundary-F1, where our method improves from 0.5061 to 0.6973. This supports the claim that the model better respects object contours rather than only matching coarse regions. On edge control, PixelControl substantially reduces Chamfer distance, showing that generated edges are not merely dense but spatially aligned with the condition.

\noindent\textbf{Visual quality.}
Fine-grained control should not degrade visual realism. Therefore, we report FID, CLIP-Img, and LPIPS for depth, segmentation, and edge control. As shown in \cref{tab:imagequality}, our method achieves the best FID in all three tasks. The improvement is especially clear for segmentation and edge control, where enforcing local structure can otherwise produce fragmented textures or over-sharpened artifacts. These results suggest that SACI and MPCL improve condition adherence while preserving the natural image prior to the pixel diffusion backbone.

\begin{table*}[t]
  \centering
  \scriptsize
  \setlength{\tabcolsep}{3pt}
  \begin{tabular*}{\textwidth}{@{\extracolsep{\fill}}lccccccccc@{}}
    \toprule
    & \multicolumn{3}{c}{Depth} & \multicolumn{3}{c}{Segmentation} & \multicolumn{3}{c}{Edge} \\
    \cmidrule(lr){2-4}\cmidrule(lr){5-7}\cmidrule(l){8-10}
    Method
    & FID \(\downarrow\) & CLIP-Img \(\uparrow\) & LPIPS \(\downarrow\)
    & FID \(\downarrow\) & CLIP-Img \(\uparrow\) & LPIPS \(\downarrow\)
    & FID \(\downarrow\) & CLIP-Img \(\uparrow\) & LPIPS \(\downarrow\) \\
    \midrule
    AnyControl & 45.8426 & 0.7280 & 0.6776 & 47.2984 & 0.7176 & 0.7049 & 48.3428 & 0.7144 & 0.6385 \\
    ControlNet & 62.9528 & 0.6850 & 0.7073 & 46.5417 & 0.7158 & 0.6910 & 46.2963 & 0.7211 & 0.6052 \\
    Ctrl-Adapter & 52.7158 & 0.6898 & 0.6920 & 41.0080 & 0.7274 & 0.6493 & 48.3437 & 0.7039 & 0.6147 \\
    OminiControl & 39.6671 & 0.7572 & 0.6283 & -- & -- & -- & 37.9785 & \textbf{0.7989} & 0.5321 \\
    PixelPonder & 34.9906 & 0.7755 & 0.5649 & -- & -- & -- & 54.5853 & 0.7076 & 0.6165 \\
    UniControlNet & 44.2812 & 0.7327 & 0.6871 & 42.6383 & 0.7376 & 0.6993 & 41.4785 & 0.7550 & 0.6249 \\
    Ours & \textbf{32.7841} & \textbf{0.7895} & \textbf{0.5318} & \textbf{29.3786} & \textbf{0.7824} & \textbf{0.5723} & \textbf{28.4531} & 0.7897 & \textbf{0.5101} \\
    \bottomrule
  \end{tabular*}
  \caption{Image quality and image consistency across depth, segmentation, and edge control. Best results are bolded.}
  \label{tab:imagequality}
\end{table*}

\noindent\textbf{Qualitative comparison.}
Visual comparisons in \cref{fig:qual} show the same trend as the quantitative results. Existing methods often recover the dominant layout but fail to preserve fine-grained condition fidelity: depth structures drift, edge controls lose or distort thin contours, and segmentation boundaries become inaccurate around local regions. In contrast, PixelControl better preserves condition-specified structures while maintaining natural image appearance. The improvement is especially visible around boundary-sensitive regions and fine structures that are easily weakened by global condition following.

\subsection{Non-large region accuracy}
\label{sec:smallobjects}

We further evaluate condition fidelity on the non-large, condition-grounded regions defined in \cref{sec:setup}.
For each selected YOLO box, we crop the corresponding region from both the input condition and the re-estimated structure map, and compute the metric within the cropped region.
For each modality, we report its primary structural metric within these regions: AbsRel for depth, Chamfer distance for edge, and Boundary-F1 for segmentation.

\begin{figure}[t]
  \centering
  \includegraphics[width=\columnwidth]{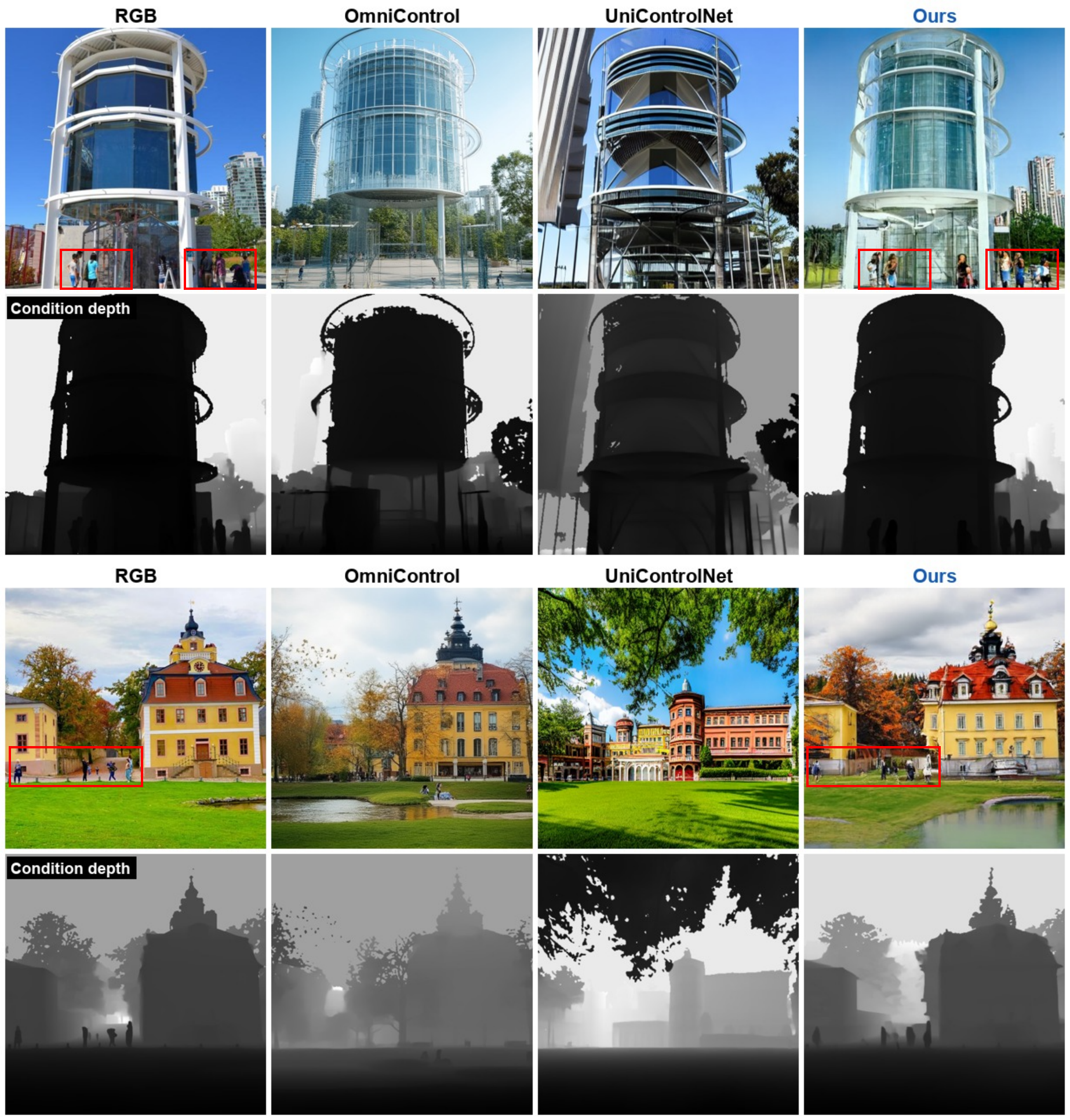}
  \caption{Non-large conditioned-region qualitative comparison. Red boxes highlight non-large objects or local structures specified by the condition. PixelControl better preserves these regions while maintaining the global structure.}
  \label{fig:nonlargequal}
\end{figure}

\cref{tab:smallobject} shows that the advantage of PixelControl becomes more pronounced on non-large conditioned objects. Compared with the best baseline, PixelControl lowers depth AbsRel from 0.2727 to 0.1419, reduces edge Chamfer from 4.381 to 2.603, and improves segmentation Boundary-F1 from 0.4728 to 0.6409. Since all evaluated regions are detector-confirmed and condition-grounded, these gains directly measure whether the model reproduces requested local objects and fine structures, rather than hallucinating plausible global layouts. The visualization in \cref{fig:nonlargequal} further illustrates this behavior: competing methods may preserve the large scene structure, but their re-estimated depth maps and generated images often miss or distort the non-large regions highlighted by the boxes.

\begin{table}[t]
  \centering
  \resizebox{\columnwidth}{!}{
  \begin{tabular}{@{}lccc@{}}
    \toprule
    Method & Depth AbsRel \(\downarrow\) & Edge Chamfer \(\downarrow\) & Seg. Boundary-F1 \(\uparrow\) \\
    \midrule
    AnyControl & 0.3430 & 6.315 & 0.4728 \\
    ControlNet & 1.4900 & 6.398 & 0.2985 \\
    Ctrl-Adapter & 1.5326 & 8.741 & 0.2986 \\
    OminiControl & 0.4757 & 7.513 & -- \\
    PixelPonder & 0.2727 & 12.568 & -- \\
    UniControlNet & 0.4303 & 4.381 & 0.2531 \\
    Ours & \textbf{0.1419} & \textbf{2.603} & \textbf{0.6409} \\
    \bottomrule
  \end{tabular}}
  \caption{Non-large conditioned-object fidelity. We evaluate only YOLO-detected non-large objects that are explicitly grounded in the condition. Best results are bolded.}
  \label{tab:smallobject}
\end{table}

\subsection{Ablation studies}
\label{sec:ablations}

We ablate the proposed components from three perspectives. First, we evaluate the overall contribution of the proposed SACI and MPCL. Second, we vary the SACI strength to study how structure-guided residual weighting affects fine details. Third, we change the MPCL scale set to verify the importance of multi-scale structural supervision. The ablations use the same evaluation protocol as the main comparison, and the PixelDiT baseline uses naive condition injection with only the FM loss.

\begin{table*}[t]
  \centering
  \resizebox{\textwidth}{!}{
  \begin{tabular}{@{}lccccccc@{}}
    \toprule
    Variant & Depth \(\downarrow\) & Seg. \(\uparrow\) & Edge \(\downarrow\) &
    Non-large Depth \(\downarrow\) & Non-large Seg. \(\uparrow\) & Non-large Edge \(\downarrow\) &
    Avg. FID \(\downarrow\) \\
    \midrule
    PixelDiT baseline & 0.2233 & 0.4824 & 12.851 & 0.3312 & 0.4218 & 6.957 & 35.82 \\
    + SACI only & 0.1814 & 0.6127 & 12.622 & 0.2184 & 0.5655 & 6.849 & 33.47 \\
    + MPCL only & 0.1666 & 0.6412 & 5.663 & 0.2019 & 0.5894 & 2.797 & 32.70 \\
    Ours & \textbf{0.1434} & \textbf{0.6973} & \textbf{5.456} & \textbf{0.1419} & \textbf{0.6409} & \textbf{2.603} & \textbf{30.21} \\
    \bottomrule
  \end{tabular}}
  \caption{
  Overall architecture ablation. The PixelDiT baseline uses naive condition injection and only the FM loss. SACI is applied only to depth and segmentation controls, so its gains mainly appear on depth and segmentation fidelity. Edge improvements are primarily brought by the pyramid cycle supervision.
  }
  \label{tab:archablation}
\end{table*}

\cref{tab:archablation} shows that PixelDiT alone is not sufficient for high-fidelity controllable generation. The PixelDiT baseline provides a useful pixel-space foundation, but naive residual injection still struggles with both full-image condition metrics and non-large objects. 
SACI mainly improves depth and segmentation fidelity because it is applied to dense or region-level conditions, whose Sobel gradients provide reliable structure weights.
Since SACI is disabled for edge control, edge metrics change only slightly after adding SACI.
MPCL substantially improves edge fidelity, especially on non-large regions, because the SoftCanny image-cycle loss directly supervises contour consistency across scales. The full model with both SACI and MPCL performs best across all metrics, indicating that the gains come from the combination of pixel-space generation, structure-aware injection, and multi-scale cycle supervision.

\begin{table}[t]
  \centering
  \resizebox{\columnwidth}{!}{
  \begin{tabular}{@{}lccc@{}}
    \toprule
    SACI strength & Depth AbsRel \(\downarrow\) & Seg. BF1 \(\uparrow\) & Avg. FID \(\downarrow\) \\
    \midrule
    No SACI (\(\alpha=0\)) & 0.2233 & 0.4824 & 35.82 \\
    \(\alpha=0.5\) & 0.2085 & 0.5564 & 34.77 \\
    \(\alpha=1.0\) & 0.1974 & \textbf{0.6211} & 33.92 \\
    \(\alpha=2.0\) & \textbf{0.1814} & 0.6127 & \textbf{33.47} \\
    \(\alpha=4.0\) & 0.1856 & 0.6041 & 33.72 \\
    \bottomrule
  \end{tabular}}
  \caption{Ablation on SACI strength. The \(\alpha\) controls how strongly structure-aware regions modulate the injected control residual.}
  \label{tab:saciablation}
\end{table}

\cref{tab:saciablation} studies the strength of the structure-aware modulation. Without SACI, control residuals are injected uniformly and the model under-emphasizes detailed regions. Increasing \(\alpha\) improves depth and segmentation fidelity until \(\alpha=2.0\), after which performance slightly drops. This suggests that overly strong modulation can over-constrain local structures and compete with semantic generation, while a moderate value gives condition-specified details enough influence during denoising.

\begin{table}[t]
  \centering
  \resizebox{\columnwidth}{!}{
  \begin{tabular}{@{}lccc@{}}
    \toprule
    MPCL setting & Depth AbsRel \(\downarrow\) & Non-large Edge \(\downarrow\) & Avg. FID \(\downarrow\) \\
    \midrule
    No MPCL & 0.1814 & 6.849 & 33.47 \\
    Low-res only (128, 64) & 0.1586 & 4.724 & 32.25 \\
    High-res only (512) & 0.1694 & 3.138 & 31.92 \\
    Full pyramid (512, 256, 128, 64) & \textbf{0.1434} & \textbf{2.603} & \textbf{30.21} \\
    \bottomrule
  \end{tabular}}
  \caption{Ablation on MPCL scales. Low-resolution scales emphasize global layout, high-resolution scales emphasize fine details, and the full pyramid provides the best balance between condition fidelity and visual quality.}
  \label{tab:mpclablation}
\end{table}

\cref{tab:mpclablation} validates the multi-scale design of MPCL. Low-resolution supervision improves global consistency but is less effective for non-large edge structures. High-resolution supervision better preserves fine details, yet it alone does not provide the same global stability or image quality. The full pyramid combines both effects and achieves the best condition fidelity and FID, supporting the design choice of verifying the generated image at multiple spatial scales.

\subsection{Multi-condition control}
\label{sec:multicond}

PixelControl also supports compositional control through modality-specific branches and a lightweight gating network over active conditions. We evaluate this ability on the joint depth+edge setting, where the model must preserve continuous scene geometry from the depth map while also following thin contours and local discontinuities from the edge map. This setting is challenging because dense geometric cues mainly constrain coarse 3D layout, whereas edge cues emphasize high-frequency structures; combining them requires the model to respect both
global shape and local boundaries.

\begin{figure}[t]
  \centering
  \includegraphics[width=\columnwidth]{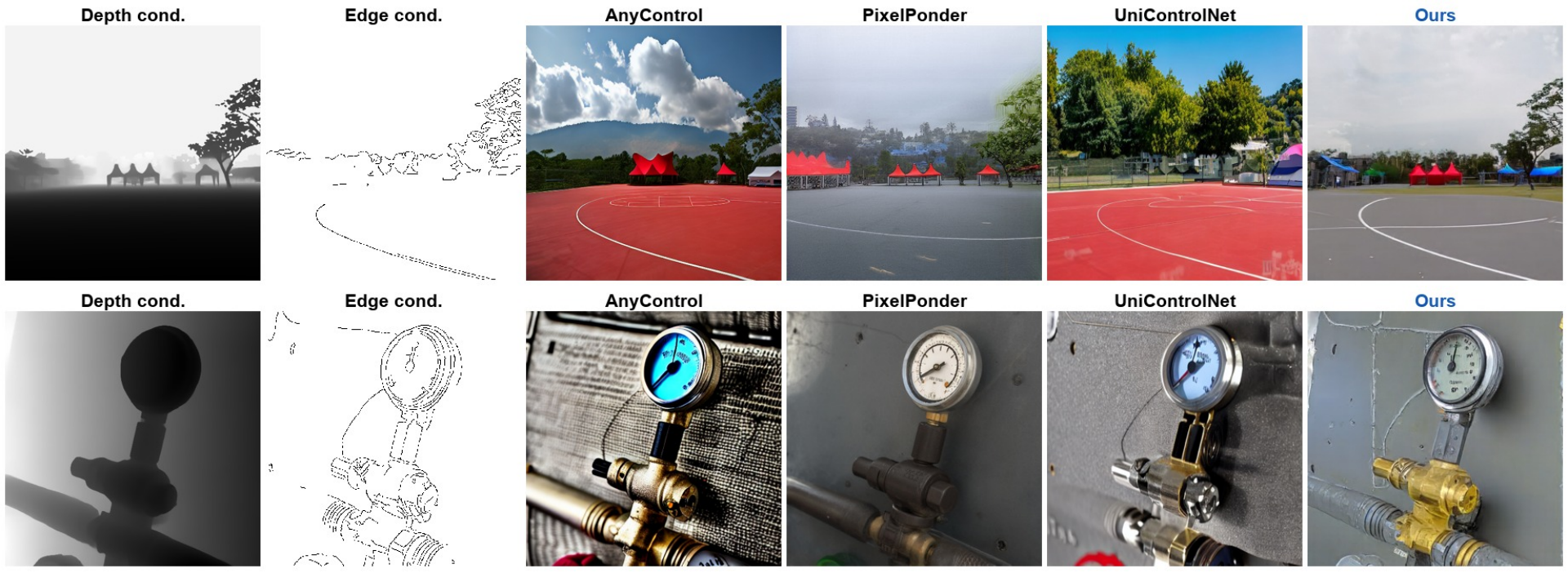}
  \caption{Depth+edge multi-condition qualitative comparison. Given both depth and edge conditions, PixelControl better preserves global geometry and local contours, producing outputs that satisfy both active controls more faithfully.}
  \label{fig:multicond}
\end{figure}

Table \ref{tab:multicond} reports depth+edge control results. PixelControl obtains the best depth AbsRel, edge Chamfer distance, and FID among methods that support compositional control.
Qualitative examples in \cref{fig:multicond} further illustrate that the model follows the depth map for global scene geometry while also respecting edge-specified contours. Competing methods tend to favor one condition, producing either plausible geometry with weak edges or sharp local details with less faithful layout.

\begin{table}[t]
  \centering
  \resizebox{\columnwidth}{!}{
  \begin{tabular}{@{}lccc@{}}
    \toprule
    Method & Depth AbsRel \(\downarrow\) & Edge Chamfer \(\downarrow\) & FID \(\downarrow\) \\
    \midrule
    AnyControl & 0.3154 & 14.82 & 49.19 \\
    OminiControl & 0.3821 & 18.06 & 42.33 \\
    UniControlNet & 0.3907 & 12.48 & 44.22 \\
    Ours & \textbf{0.1822} & \textbf{6.42} & \textbf{33.27} \\
    \bottomrule
  \end{tabular}}
  \caption{Depth+edge multi-condition control. We evaluate structural fidelity for both conditions and report FID for visual quality.}
  \label{tab:multicond}
\end{table}

\section{Conclusion}
\label{sec:conclusion}

We presented PixelControl, a pixel-space controllable text-to-image diffusion framework for fine-grained spatial condition fidelity.
PixelControl introduces Structure-Aware Control Injection to amplify control residuals around condition boundaries and discontinuities, and Multi-Scale Pyramid Cycle Loss to enforce structural consistency from global layout to local detail.
It supports depth, segmentation, edge, and their combinations through modality-specific branches with lightweight layer-wise gating.
Experiments show that PixelControl consistently improves both structural fidelity and visual quality across multiple condition types, with especially clear gains on boundary alignment and non-large conditioned regions.

{
    \small
    \bibliographystyle{ieeenat_fullname}
    \bibliography{main}
}

\end{document}